\documentclass[sigconf,nonacm]{acmart}
\AtBeginDocument{%
  }

\setcopyright{acmlicensed}
\copyrightyear{2018}
\acmYear{2018}
\acmDOI{XXXXXXX.XXXXXXX}
\acmConference[Conference acronym 'XX]{Make sure to enter the correct
  conference title from your rights confirmation email}{June 03--05,
  2018}{Woodstock, NY}
\acmISBN{978-1-4503-XXXX-X/2018/06}

\begin{document}

\title{A Vision for Geo-Temporal Deep Research Systems}
\subtitle{Towards Comprehensive, Transparent, and Reproducible Geo-Temporal Information Synthesis}

\author{Bruno Martins}
\email{bruno.g.martins@tecnico.ulisboa.pt}
\affiliation{%
  \institution{Instituto Superior Técnico and INESC-ID\\Univesrity of Lisbon}
  \city{Lisbon}
  \country{Portugal}}

\author{Piotr Szymański}
\email{piotr.szymanski@pwr.edu.pl}
\author{Piotr Gramacki}
\email{piotr.gramacki@pwr.edu.pl}
\affiliation{%
  \institution{Wrocław University of Science and Technology}
  \city{Wrocław}
  \country{Poland}
}

\renewcommand{\shortauthors}{Martins et al.}

\begin{abstract}
The emergence of Large Language Models (LLMs) has transformed information access, with current LLMs also powering deep research systems that can generate comprehensive report-style answers, through planned iterative search, retrieval, and reasoning. Still, current deep research systems lack the geo-temporal capabilities that are essential for answering context-rich questions involving geographic and/or temporal constraints, frequently occurring in domains like public health, environmental science, or socio-economic analysis. This paper reports our vision towards next generation systems, identifying important technical, infrastructural, and evaluative challenges in integrating geo-temporal reasoning into deep research pipelines. We argue for augmenting retrieval and synthesis processes with the ability to handle geo-temporal constraints, supported by open and reproducible infrastructures and rigorous evaluation protocols. Our vision outlines a path towards more advanced and geo-temporally aware deep research systems, of potential impact to the future of AI-driven information access.
\end{abstract}


\begin{teaserfigure}
  \centering
  \includegraphics[width=0.9\textwidth]{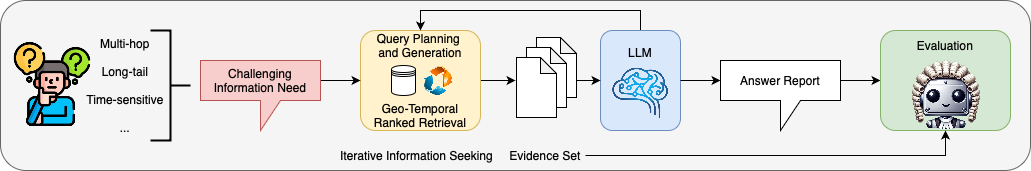}
  \vspace{-0.2cm}
  \caption{Overview on the possible architecture of a geo-temporal deep research system.}
  \Description{Overview on the possible architecture of a geo-temporal deep research system.}
  \label{fig:teaser}
\end{teaserfigure}


\maketitle

\vspace{-0.25cm}
\section{Introduction}

Large Language Models (LLMs) have dramatically transformed the landscape of information access, enabling systems that transcend traditional ranked retrieval and instead generate comprehensive, report-style answers to complex, multi-faceted queries. These deep research systems, exemplified by recent commercial developments such as OpenAI, Gemini, and Perplexity's deep research modes, as well as several open-source replicas (e.g., GPT Researcher, Open Deep Research, etc.) and similar academic endeavors~\cite{li2025search,jin2025search,jin2025empirical,url16,zheng2025deepresearcher,song2025r1,li2025r3}, integrate iterative search, retrieval, and reasoning over vast textual corpora to address intricate user needs. With deep research pipelines supporting recent developments such as artificial scientists capable of generating top-tier conference papers~\cite{zochi2025,lu2024ai,yamada2025ai,tang2025ai,team2025novelseek,schmidgall2025agent}, and as industry leaders like Google and Apple signal a strategic shift towards AI-driven search and research solutions, it is clear that the future of information access lies in increasingly autonomous systems, capable of navigating and synthesizing large-scale data for complex information needs. 

While deep research systems have already demonstrated their potential, a critical frontier remains underexplored: the integration of geo-temporal reasoning into deep research pipelines. We argue that incorporating the geographic and temporal dimensions into information access is essential for answering complex, contextually rich questions that span space and time~\cite{piryani2025s,url1}, which are commonly involved in producing actionable insights relevant to a broad spectrum of disciplines, from public health and environmental science to economics and urban planning. For instance, queries may seek regulatory changes across regions, environmental trends over decades, or health policy impacts in specific communities. However, extending existing deep research systems to support geo-temporal constraints introduces substantial new challenges.

By design, deep research systems are agentic, relying on planned iterative retrieval and reasoning over large text datasets (e.g., LLMs are used for planning, exploration, and tool calling, iteratively diving deeper into subtopics while maintaining a comprehensive view of the research subject). To enable geo-temporal reasoning, these processes must be augmented with techniques for geo-temporal text analysis and retrieval~\cite{url1,url2,url4,url5,piryani2025s,yu2025spatial}, for instance including query planning and generation strategies that account for constraints in terms of place names, geographic footprints expressed through points or polygonal boundaries, calendar dates or time intervals, and ranked retrieval methods that account with proximity and/or containment relations~\cite{url2,url4,url5}, as well as with the need for result diversity across the geo-temporal dimensions~\cite{url7,url9}. The necessarily limited window of information that can be effectively processed by LLMs further constrains these systems, making it imperative for retrieval mechanisms to consider the salience and relevance of documents with respect to geo-temporal specificity. Moreover, accurate synthesis of retrieved information depends on precise parsing and disambiguation of geo-temporal cues in text~\cite{zhang2024survey,url6,url13,gautam2024discourse}, potentially requiring external tools such as geo-coders, gazetteers, and mapping services to accurately ground information into real-world contexts, given that LLMs alone are likely to produce many errors in these tasks~\cite{abbasi2025world}.

Moreover, developing geo-temporal deep research systems is not solely a technical challenge, but also an infrastructural one. For instance, current research in deep information retrieval and synthesis is constrained by the reliance on commercial Web search APIs, whose proprietary nature limits transparency, and where search result evolution undermines reproducibility~\cite{coelho2025deepresearchgym}. This hinders progress, as new developments require access to large-scale, stable, and openly accessible platforms, capable of supporting comprehensive retrieval experiments. To support advancements, the research community must invest in building environments with broad and diverse coverage, i.e. search infrastructures that (a) simulate real-world conditions, (b) index large Web corpora, (c) incorporate advanced retrieval models supporting fine-grained queries and ranked retrieval involving geo-temporal constraints, and (d) incorporate geo-temporal modules for specialized tasks such as geo-coding and reverse geo-coding, map routing, or the computation of geo-spatial statistics, aggregations, and interpolations. Such environments would empower researchers to audit system behavior, analyze evidence influence, and re-run experiments under consistent conditions, ensuring fair benchmarking and reproducibility.

Equally crucial is the development of rigorous evaluation protocols tailored to geo-temporal deep research, aimed at assessing progress in tasks that entail the extraction and analysis of geo-temporally constrained data from diverse sources, including policy documents, scientific literature, or general Web content. For traditional retrieval augmented generation systems, and even for deep research, some benchmarks have already been established as testbeds for assessing retrieval effectiveness and response quality in complex high-engagement queries~\cite{url16,url17,coelho2025deepresearchgym,url17,krishna2024fact,thakur2025support,ru2024ragchecker,saad2023ares,sivasothy2024ragprobe,mialon2023gaia,phan2025humanity,pradeep2025ragnarok}. These existing benchmarks can inform the creation of new ones, and previous methodologies, including LLM-as-a-Judge approaches~\cite{gu2024survey,li2024generation,li2024llms}, can perhaps also be extended to evaluate systems' capacity to align with geo-temporal constraints, assessing not only factual grounding and report quality, but also spatial and temporal relevance, coherence, and diversity.

This paper presents a vision towards advanced deep research systems with geo-temporal capabilities - see Figure~1 for a high-level architectural overview. Research towards this vision must address the technical, infrastructural, and evaluative challenges inherent in integrating geo-temporal reasoning into deep research pipelines. One such research program is also interdisciplinary in nature, as previous developments in geographical~\cite{url1,url2,zhang2024survey,url6,url13,yu2025spatial} and temporal~\cite{url3,url4,url5,piryani2025s,url3,gautam2024discourse} text analysis and retrieval, and geo-temporal knowledge representation and reasoning~\cite{url13,url14,url10,url15,lin2023geogalactica,deng2024k2,van2025opportunities}, can now be combined with recent LLM advances. We call for the development of open, reproducible platforms and methodologies that enable meaningful progress in this exciting field, with the potential for transforming how we access, synthesize, and act on information in a world defined by geo-temporal interdependencies.

\vspace{-0.25cm}
\section{Towards Geo-Temporal Information Synthesis}

This section outlines our vision for deep research systems enhanced with geo-temporal capabilities. Section 2.1 explores the broad and complex spectrum of questions that such systems must be equipped to address. Section 2.2 discusses the essential building blocks required to develop these systems. Finally, Section 2.3 highlights the critical considerations for designing open evaluation protocols that can accurately measure system effectiveness and reliability.

\vspace{-0.25cm}
\subsection{Complex Geo-Temporal Information Needs}

Geo-temporal deep research systems have the potential to address a diverse range of complex information needs, combining iterative retrieval, reasoning, and synthesis across corpora. For example, they can be used to uncover geo-temporal correlations between environmental factors and human health outcomes, track the evolution of regulatory and policy landscapes over time and across regions, and provide comprehensive analyses of disaster and crisis events as they unfold. Additionally, they can explore dynamic phenomena such as epidemiological trends, urban development patterns, economic changes, environmental shifts, and demographic transformations. By integrating geo-temporal reasoning into retrieval and synthesis pipelines, these systems have the potential to enable nuanced insights that are critical for decision-making across disciplines.

Crucially, the focus of geo-temporal deep research is not on automating GIS-style spatial analysis -- including code generation or tool execution for complex spatial computations -- but rather on orchestrating processes for iterative retrieval, reasoning, and synthesis from vast collections of unstructured textual documents. The goal is to produce comprehensive, report-style answers that integrate diverse forms of evidence. Even though geo-temporal analysis is required, and some cases can consider structured tabular outputs (e.g., summarizing data by region or time period), the focus is on text analysis for producing reports, e.g. which can later be incorporated into external GIS workflows, if needed. The emphasis on text-based synthesis distinguishes the envisioned systems from previous efforts aimed at automating GIS workflows~\cite{url8,url11,zhang_geogpt_2024}, although both types of approaches can indeed be combined. 

A list of question types, illustrating the breadth and complexity of the information needs that can be supported, is given next:

\begin{itemize}
\item {\bf Multi-hop:} Queries requiring information synthesis from multiple sources to construct a complete answer (e.g., {\it what policies have been implemented in response to air pollution incidents in European urban centers over the last decade, and how have these policies affected public health outcomes?}).

\item {\bf Long-tail:} Questions about obscure facts, rare entities, or low-resource domains with limited coverage (e.g., {\it what is known about the economic impact of a small rural festival in a specific region of Northern Europe?}).

\item {\bf Time-sensitive:} Questions involving explicit or implicit time anchors and/or temporal constraints, likely also involving the disambiguation of vague temporal expressions (e.g., {\it what were the immediate policy responses to the COVID-19 outbreak in different European countries in early 2020?}).

\item {\bf Location-sensitive:} Questions involving explicit or implicit geospatial anchors, likely requiring the handling of vague references to specific regions, cities, or landmarks, and likely also requiring reasoning about containment or proximity between locations (e.g., {\it what environmental policies have been enacted near the Amazon rainforest in the last five years?}).

\item {\bf Freshness-sensitive:} Real-time queries needing recent information, likely requiring real-time retrieval from sources like news or social media (e.g., {\it what are the most recent developments in wildfire containment efforts in California?}).

\item {\bf Diversity-sensitive:} Queries seeking a breadth of perspectives, covering geographic, temporal, cultural, or thematic diversity (e.g., {\it what are the different community responses to renewable energy projects across Europe?}).

\item {\bf Distracting information:} Queries susceptible to noise or false data, due to their focus on ambiguous or controversial topics (e.g., {\it what are the current theories about the origins of a controversial archaeological site, and how do they differ?}).

\item {\bf False premise:} Questions based on incorrect assumptions, requiring systems to detect and correct the underlying misunderstandings (e.g., {\it what are the current prices of high-end CPUs and/or GPUs manufactured in Europe?}).

\item {\bf Temporal comparison:} Queries asking for comparisons across different time periods (e.g., {\it how have hurricane frequencies in the Caribbean changed over the past 50 years?}).

\item {\bf Geographical comparison:} Queries asking for comparisons across different places (e.g., {\it how do air quality levels compare between urban and rural areas in Europe?}).

\item {\bf Temporal aggregation:} Queries that necessitate aggregating information across time (e.g., {\it what is the cumulative economic loss, for different southern European countries, caused by natural disasters in the last decade?}).

\item {\bf Geographical aggregation:} Queries that require aggregating information across space (e.g., {\it how many endangered species exist across phyla and geographical continents?}).

\item {\bf Granularity-sensitive:} Queries that require reasoning over different levels of spatial or temporal granularity, likely involving nested or hierarchical relationships (e.g., {\it what trends emerge in European employment rates, when comparing monthly vs. quarterly data at regional and national levels?}).

\item {\bf Temporal ordering:} Queries about the sequence or order of events over time (e.g., {\it what events led up to the signing of a major environmental treaty?}).
\end{itemize}

Notice that most real-world information needs are inherently multi-faceted, involving complex constraints that span multiple dimensions -- temporal, geographic, and thematic. Even the example questions provided in the previous list frequently combine these elements, illustrating the need for integrated reasoning across diverse contexts. For instance, a query like {\it what were the key differences in health policy decisions made by coastal cities during the early stages of the COVID-19 pandemic, and how did these decisions evolve over the following year} exemplifies the overall complexity, requiring temporal tracking, geographical specificity, and thematic depth. 

The range of question types that was given as example not only underscores the breadth of information needs these systems must support, but also hints at the architectural components that are essential for such systems -- including capabilities for planned iterative retrieval, geo-temporal reasoning, and evidence synthesis -- as well as the evaluation protocols necessary to assess effectiveness. These aspects will be further explored in the next sub-sections.

\vspace{-0.25cm}
\subsection{Architectural Components}

Designing effective geo-temporal deep research systems introduces a range of architectural challenges that go beyond those encountered in standard deep research frameworks. At the heart of these challenges lies the need to handle geo-temporal constraints in user queries, necessitating careful consideration in the design of components for query generation, initial evidence retrieval, re-ranking, and information synthesis. Figure 1 provides a general illustration for the main components involved in one such system, and as most components would likely involve neural models, the entire system (or some of the key components) can be adjusted end-to-end through reinforcement learning, using task-specific evaluation metrics as reward functions for training~\cite{zheng2025deepresearcher,sun2025simpledeepsearcher,song2025r1,li2025r3}.

Query generation plays a pivotal role in translating information needs into actionable search queries. Unlike standard systems that plan and formulate queries primarily with a topical focus, geo-temporal deep research systems must construct queries that explicitly capture geo-temporal constraints, such as particular locations, regions, dates, or time periods of interest. This includes generating queries that seek not only geo-temporally localized information, but also diverse results spanning multiple spatial regions or temporal windows. Techniques from Geographic Information Retrieval (GIR) and Temporal Information Retrieval (TIR) provide inspiration here~\cite{url1,piryani2025s}, offering strategies for extracting and integrating geo-temporal signals from natural language queries.

The retrieval and re-ranking components also require adaptations to deal with geo-temporal queries~\cite{url1,url2,url4,url5,piryani2025s,yu2025spatial}, and at the same time they must operate under constraints of relevance, precision, and scalability, particularly as LLMs can only process a limited number of documents to effectively summarize content and synthesize reports. This requires highly accurate retrieval pipelines that can identify and prioritize the most relevant documents. Here, instruction-following and reasoning-based retrieval models offer promising directions~\cite{zhuang2025towards,zhou2024beyond,weller2024followir,das2025rader,shao2025reasonir,zhang2025rearank}, combining the structured handling of geo-temporal constraints, as in GIR/TIR systems, with the flexibility and task-following capabilities of advanced LLMs. These approaches can better interpret complex queries (e.g., those requiring reasoning about proximity, containment, or temporal ordering), and identify small and high-quality evidence sets.

A significant limitation in current deep research systems arises from infrastructure dependencies~\cite{coelho2025deepresearchgym}. Most state-of-the-art systems rely on commercial Web search APIs to perform retrieval, which introduces critical challenges. These APIs are opaque, hindering transparency in the retrieval process, and their evolving nature undermines reproducibility and fair benchmarking. Geo-temporal deep research systems, also due to their specialized retrieval needs, further motivate the need for dedicated infrastructures. These should consider document collections with broad and diverse coverage, embedding models fine-tuned for effective geo-temporal retrieval, and scalable nearest-neighbor search indices that can efficiently handle complex, high-dimensional retrieval tasks. Developing such infrastructures in the open research ecosystem is essential for progress.

Besides query planning and document retrieval, the summarization and reasoning components also involve additional considerations to synthesize information across multiple documents, in a way that incorporates the geo-temporal context~\cite{yu2025spatial,url3,url14}. This involves not only integrating retrieved content, but also accurately interpreting and disambiguating geo-temporal information~\cite{zhang2024survey,url6,url13,url12,gautam2024discourse}, such as place names, addresses, temporal expressions, and complex spatial relationships. LLMs used for summarization may require multiple processing steps, coordinating tasks such as geo-coding, reverse geo-coding, map-based reasoning, and even performing lightweight GIS operations—like aggregating counts or interpolating values over spatial regions. These capabilities can be improved and further augmented by integrating external tools and services (e.g., gazetteers, geo-coding APIs, routing services, or temporal reasoning libraries) to support the generation of precise and grounded responses. Again, previous developments in GIR/TIR can inform component design~\cite{url1,piryani2025s,yu2025spatial,url13,url14}.

Overall, the architecture of geo-temporal deep research systems must harmonize sophisticated retrieval pipelines, stable and transparent infrastructures, and advanced reasoning components. This orchestration supports the generation of long-form, spatio-temporally aware reports that address complex information needs, a capability well beyond what current systems can achieve.

\vspace{-0.25cm}
\subsection{Challenges for Evaluation}

Besides properly evaluating individual components, advancing geo-temporal deep research requires comprehensive evaluation protocols that can assess overall performance under realistic scenarios. While previous efforts, e.g. the TREC RAG shared task~\cite{pradeep2025ragnarok,thakur2025support,pradeep2024initial} and others~\cite{url17,krishna2024fact,ru2024ragchecker,saad2023ares,sivasothy2024ragprobe,mialon2023gaia,phan2025humanity}, have provided valuable baselines for the evaluation of Retrieval-Augmented Generation (RAG) systems, these approaches fall short of addressing the unique challenges introduced by deep research. RAG typically operates over static domain-specific corpora and is evaluated using standard retrieval scores (e.g., NDCG) and text generation metrics based on the overlap with ground-truth answers. Deep research, on the other hand, is characterized by interactive agentic workflows over large-scale collections, including the ever-changing Web, which introduces challenges not only for system design but also in defining fair, reproducible, and meaningful evaluation methodologies~\cite{url16,url17,coelho2025deepresearchgym}.

One of the central challenges lies in the nature of the outputs generated by deep research systems, which are long-form, report-style answers that integrate information from multiple sources, often involving multi-step reasoning. Evaluating such outputs goes well beyond simple fact-checking or retrieval relevance assessments. Metrics like FActScore~\cite{min2023factscore} or Key Point Recall~\cite{qi2024long} offer a partial solution, e.g. by measuring the coverage of salient points from retrieved sources in long-form outputs. However, even these metrics are limited in scope, especially when considering geo-temporal dimensions. Alternatively, methods like LLM-as-a-Judge~\cite{gu2024survey,li2024generation,li2024llms}, which use language models to assess the quality, relevance, and faithfulness of generated outputs, can be extended for deep research. Yet, all these methods need adaptations to account for geo-temporal relevance, coherence, and diversity -- criteria that are essential for real-world information needs involving geo-temporal reasoning.

Building on appropriate metrics, benchmarks also play a crucial role in shaping evaluation. Datasets like researchy questions~\cite{url17} or InfoDeepSeek~\cite{url16} can provide a starting point for building benchmarks with complex high-engagement queries. However, these datasets need to be adapted to focus specifically on geo-temporal constraints, either by filtering the questions to retain those with explicit geo-temporal elements, or by creating new questions reflecting these dimensions. Inspiration can also be drawn from datasets designed for assessing GIScience-related capabilities in LLMs~\cite{url10,url15,lin2023geogalactica,deng2024k2,van2025opportunities}, although these primarily focus on short-answer queries, underscoring the need for datasets that address the complexities of long-form, geo-temporally grounded outputs.

The reproducibility challenge further complicates evaluation. Because deep research systems often retrieve information from dynamic Web collections, it becomes difficult to ensure consistent conditions across tests~\cite{coelho2025deepresearchgym}. Evaluation protocols must therefore be supported by retrieval infrastructures that offer stable, broad, and diverse coverage, and that can faithfully simulate real-world settings. Without this, it is nearly impossible to conduct fair comparisons of system performance, or to track progress over time.

Ultimately, evaluating geo-temporal deep research systems requires a multi-faceted approach. This involves adapting existing methodologies, e.g LLM-as-a-Judge~\cite{gu2024survey,li2024generation,li2024llms}, to incorporate assessments of geo-temporal relevance, coherence, and diversity. It also demands the construction of new, open benchmarks that represent complex, real-world information needs. Only through rigorous, comprehensive, and reproducible evaluation protocols can we ensure that these systems are not only technically capable, but also effective and trustworthy in addressing the increasingly complex geo-temporal information needs of modern users.

\vspace{-0.25cm}
\section{Conclusions}

We outlined a vision towards deep research systems that incorporate geo-temporal reasoning capabilities. By examining the unique challenges associated with geo-temporal information access, including the required component specializations as well as infrastructural requirements for open and reproducible systems, we have highlighted the need for a rich research agenda. Furthermore, we emphasize the importance of rigorous evaluation protocols that assess not just factual accuracy and synthesis quality, but also geo-temporal relevance and diversity. The integration of geo-temporal reasoning into deep research systems offers transformative opportunities across a wide range of disciplines, and future work towards this vision must focus on building open infrastructures, advancing geo-temporal retrieval and reasoning techniques, and developing evaluation frameworks to guide and measure progress.

\begin{acks}
We thank 
the colleagues with whom we have discussed these ideas
for their valuable comments and suggestions on a previous version of this manuscript. 
Bruno was supported by the Portuguese Recovery and Resilience Plan through project C645008882-00000055 (i.e., the Center For Responsible AI), and by Fundação para a Ciência e Tecnologia (FCT) through project UIDB/50021/2020 (DOI:10.54499/UIDB/50021/2020).
\end{acks}

\bibliographystyle{ACM-Reference-Format}
\bibliography{sample-base}


\end{document}